\documentclass[11pt]{article}

\usepackage[utf8]{inputenc}
\usepackage[T1]{fontenc}
\usepackage[english]{babel}
\usepackage{csquotes}

\usepackage{lmodern}
\usepackage{microtype}
\usepackage[scaled]{helvet}
\usepackage{eulervm}
\normalfont

\usepackage[a4paper, margin=1.15in]{geometry}

\usepackage[blocks, affil-it]{authblk}

\usepackage{color}
\usepackage{graphicx}
\usepackage{tikz}

\definecolor{darkgreen}{RGB}{33, 97, 11}
\definecolor{darkblue}{RGB}{11, 56, 97}
\definecolor{darkred}{RGB}{97, 11, 11}
\definecolor{lightgray}{gray}{0.7}

\usepackage{amsmath}
\usepackage{amssymb}
\usepackage{stmaryrd}
\usepackage{amsthm}
\usepackage{amsfonts}

\theoremstyle{plain}

\theoremstyle{definition}

\usepackage{enumitem}
\setlist{noitemsep}

\usepackage[
	unicode=true,
	pdfborder={0 0 0},
	colorlinks=true,
	linkcolor=darkblue,
	citecolor=darkgreen,
	filecolor=darkred,
	urlcolor=darkred,
	pdftitle={Cognitive Thermometers: Machine Learning and Logical Complexity},
	pdfauthor={Shane Steinert-Threlkeld and Jakub Szymanik},
	pdfsubject={Semantic universals, learnability, logical complexity},
	pdfkeywords={semantic universals, monotonicity, quantifiers, learnability,
	             logical complexity, language of thought, neural networks},
]{hyperref}

\usepackage[
	natbib=true,
	style=authoryear,
	backend=biber,
	useprefix=true    
]{biblatex}

\DeclareSortingNamekeyTemplate{
  \keypart{\namepart{family}}
  \keypart{\namepart{prefix}}
  \keypart{\namepart{given}}
  \keypart{\namepart{suffix}}
}
\renewbibmacro{begentry}{\midsentence}

\usepackage{xurl}

\renewenvironment{abstract}{%
	\medskip\noindent\textbf{Abstract}~~\small
}{%
	\medskip
}

\begin{document}

\title{Cognitive Thermometers:\\ Machine Learning and Logical Complexity%
\thanks{To appear in \emph{Logic \& AI}, edited by Johan van Benthem,
Levin Hornischer, Thomas Icard and Fenrong Liu (Stanford, CA: CSLI
Publications), which will be published open access. This is the
authors' preprint; please cite the book version once it is available.
Comments welcome.}}

\author[1,*]{Shane Steinert-Threlkeld}
\author[2,*]{Jakub Szymanik}
\affil[1]{University of Washington, Department of Linguistics}
\affil[2]{University of Trento, Center for Mind/Brain Sciences and Department of Information Engineering and Computer Science}
\affil[*]{Authors listed alphabetically}

\date{\today}

\maketitle

\begin{abstract}
How does the human mind represent semantic categories? Why do
natural languages favor certain meanings over others? Prior explanations have
relied on logical definability and complexity, but these are highly sensitive to
the choice of logical language, rendering some design choices unmotivated.
In this article, we propose that machine learning provides a somewhat more
agnostic approach to measuring semantic complexity. We review emerging
evidence that logic and machine learning often yield converging results on
relative complexity and its resulting effects in semantic typology.
Where they diverge, learning appears to be a better explanation than logical complexity.
We argue that treating machine
learning models as ``cognitive thermometers'' enables a unified approach to
complexity that bridges symbolic logic and connectionist AI.
\end{abstract}

\section{Ease of Learning and Semantic Universals}
\label{sec:ease-learning}

The natural languages of the world exhibit tremendous variation at every level, from sound (phonetics and phonology) to grammar (morphology and syntax) to meaning (semantics).  Famously, however, this variation is \emph{constrained}: despite their differences, languages also exhibit a great deal of shared structure.  A central concern of theoretical linguistics is to explain this constrained variation: what, out of the large space of mathematically possible linguistic systems, distinguishes the humanly possible ones?

When it comes to semantics, a prominent line of work has argued that \emph{ease of learning} plays a central role \parencite{steinert-threlkeldLearnabilitySemanticUniversals2019, steinert-threlkeldExplanationVeridicalUniformity2020, steinert-threlkeldEaseLearningExplains2020, douvenLearnabilityNaturalConceptsforthcoming, maldonadoLearnabilityConstraintsSemantics2022, chemlaConnectingContentLogical2019}.  In this domain, the constrained variation often takes the form of lexical property universals, which state that all expressions of a particular kind express meanings with a particular property \parencite{steinert-threlkeldComputationalExplanationsSemanticinprep}.  For example (developed more in the next section), it has been argued that all determiners (words like `some', `most', and the like) denote \emph{monotone} quantifiers. The so-called \textsc{Ease of Learning Hypothesis} then states that meanings with these distinguished properties are easier to learn than meanings without those properties; this can explain the prevalence of these semantic universals by additionally assuming that languages lexicalize easier-to-learn meanings and rely on grammar and compositional interpretation for more complex ones.  For example: if the hypothesis holds, monotone quantifiers would be easier to learn than non-monotone ones.

To argue for this hypothesis, we use machine learning as a kind of \emph{cognitive thermometer}.  In a given empirical domain, the goal is to measure the `temperature' of all mathematically possible meanings.  One can think of `warmer' meanings as the easier-to-learn ones and `cool' ones as harder.  Once this `heatmap' on the space of possible meanings has been made, one hopes to find that natural language meanings lie in the warmer regions of the space.  The use of machine learning both enables the measurement of ease-of-learning at a scale that is not feasible with human learners,\footnote{For human learning evidence complementing the machine learning experiments in the next section, see \cite{RAMOTOWSKA2026106523} who show in an artificial-language-learning study that adults' acquisition of a novel quantifier tracks the same universals, including monotonicity, described below for neural learners.} while simultaneously avoiding confounds that come from the fact that people have already acquired languages which satisfy given semantic universals.

In the next section, we illustrate this approach with one prominent, concrete example: the monotonicity of quantifiers.  From there, we discuss the relationship between this machine learning approach and other forms of complexity, including logical complexity, before discussing future directions.

\section{Example: Monotonicity of Quantifiers}
\label{sec:examples}

In the linguistic context, monotonicity concerns reasoning via predicate replacement, typically preserving entailment between expressions when one predicate is replaced with a more general one (upward monotonicity) or a more specific one (downward monotonicity) \parencite{VanBenthem1986, Benthem1991, icardiiiRecentProgressMonotonicity2014}.  These properties form a lexical property universal when looking at quantifiers, the meanings of expressions like \emph{some}, \emph{all}, \emph{most}, and their cross-linguistic ilk.  For example, from the sentence `Most Seattleites like to go hiking.', it follows that `Most Seattleites like to go outdoors.', where all we have done is replace a more specific phrase (`go hiking') with a more general one (`go outdoors').

In the application of the logical theory of generalized quantifiers \parencite{Mostowski1957, Lindstrom1966} to natural language pioneered by \textcite{barwiseGeneralizedQuantifiersNatural1981}, this property can be cast as the monotonicity of the characteristic function of a generalized quantifier.  The latter authors proposed a lexical property universal: all determiners (the relevant syntactic category) denote monotone quantifiers.\footnote{For reasons that do not concern us presently, their official formulation also allowed conjunctions of monotone quantifiers.  For more detailed treatments, see \cite{petersQuantifiersLanguageLogic2006, Szymanik:2015kq}.}

\begin{figure}[ht]
    \centering
    \includegraphics[width=0.48\textwidth]{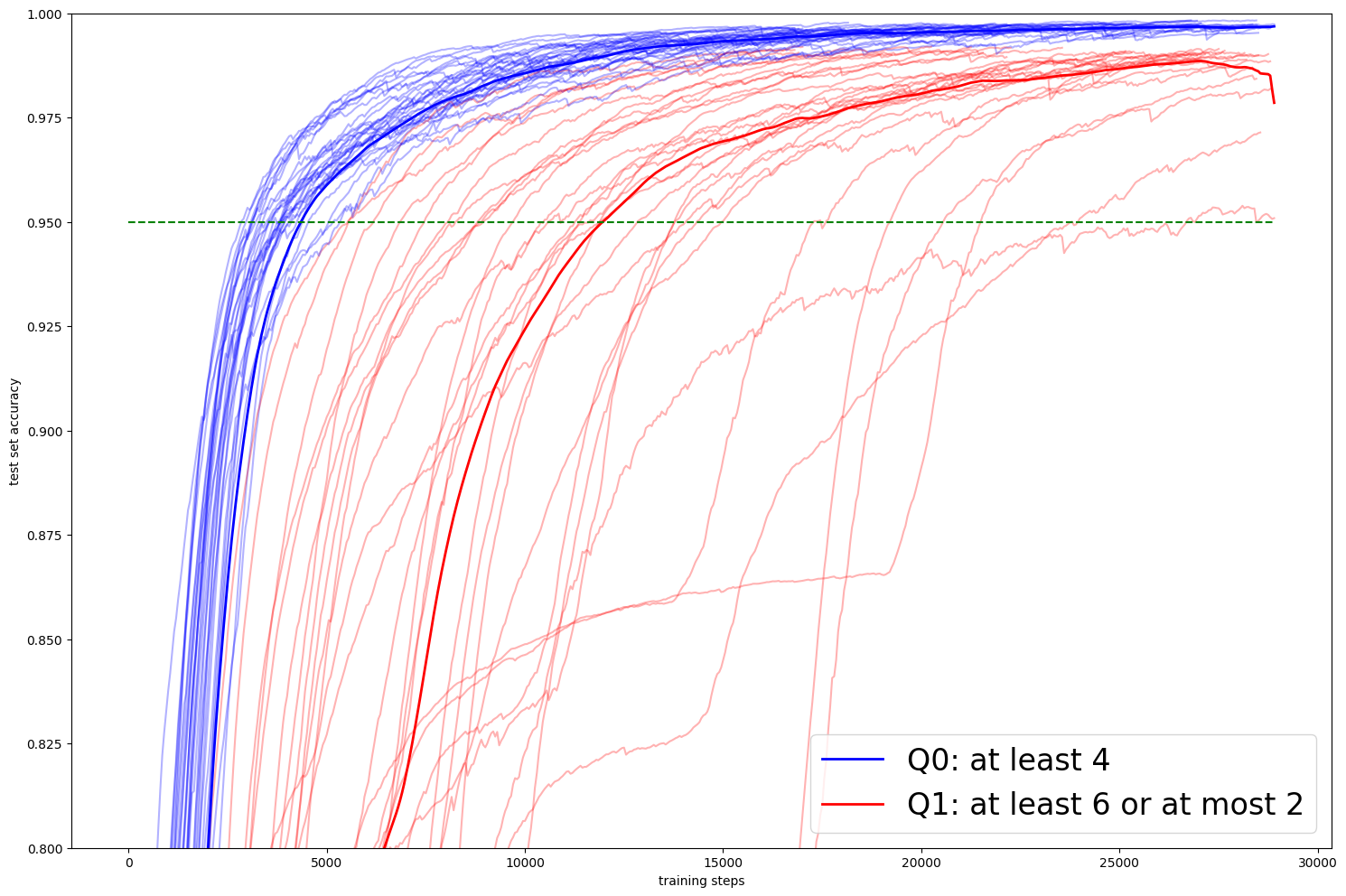}
    \includegraphics[width=0.48\textwidth]{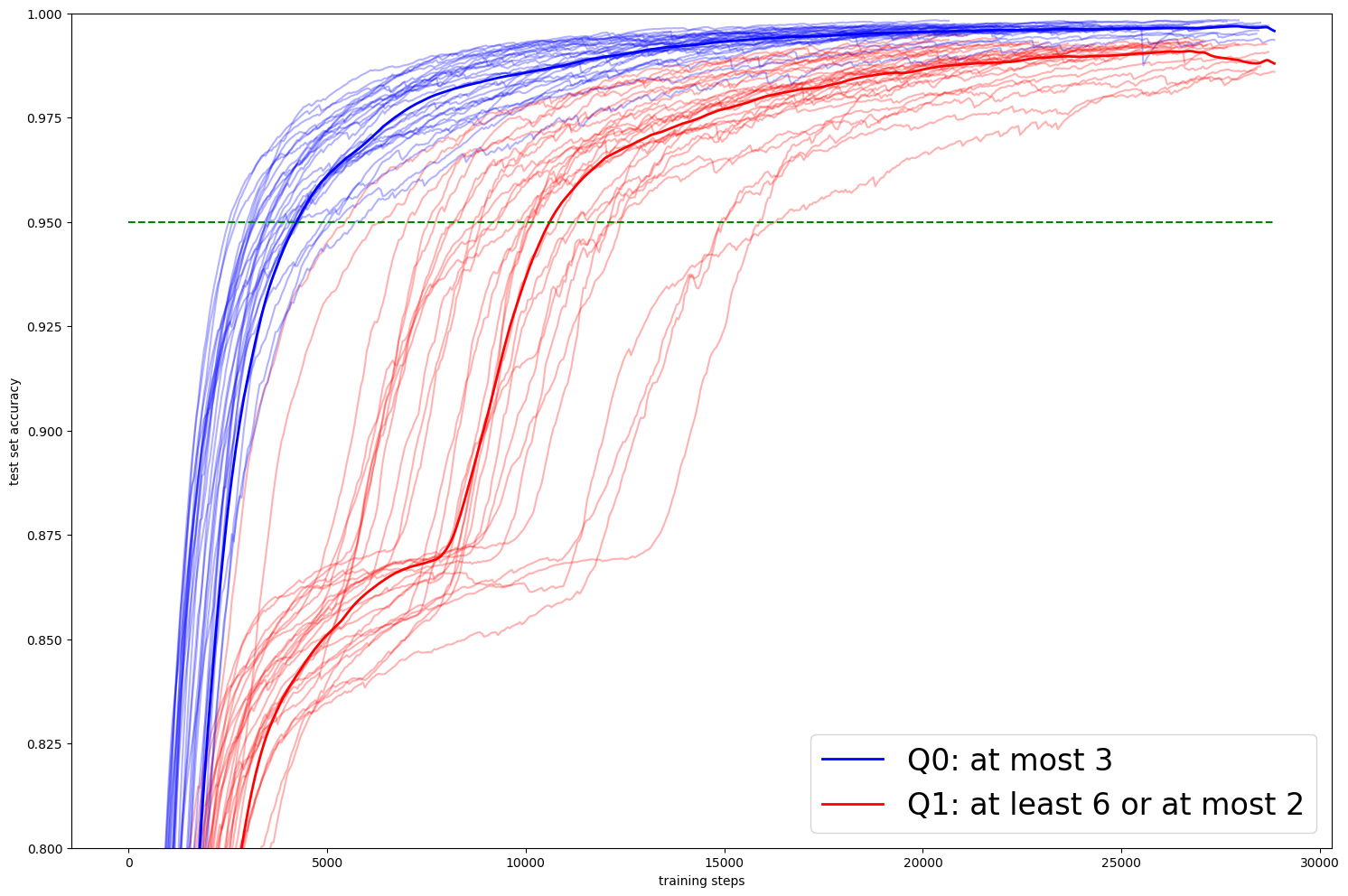}
    \caption{Results from two experiments in \cite{steinert-threlkeldLearnabilitySemanticUniversals2019}.  The $x$-axis shows the number of training steps.  The $y$-axis shows the neural models' accuracy on instances that were not seen during training (to test whether the model has truly extracted the meaning and not simply memorized examples).  Blue is a monotone quantifier (\emph{at least 4} in the left, \emph{at most 3} on the right) and red a non-monotone one (\emph{at least 6 or at most 2} in both).}
    \label{fig:monotonicity-pairs}
\end{figure}

As a first test of the \textsc{Ease of Learning Hypothesis}, \textcite{steinert-threlkeldLearnabilitySemanticUniversals2019} trained neural networks to learn the meaning of quantifiers in a minimal pair setting, comparing the upward monotone \emph{at least 4} and downward monotone \emph{at most 3} with the non-monotone \emph{at least 6 or at most 2}.  These models take as input a representation of a model-theoretic model, a label (akin to a nonce word) for a quantifier, and are trained to predict whether the quantifier is true or false in that situation.  We experimentally manipulated the meaning of the nonce word, i.e.\ what the gold mapping to truth values was. The results of these two experiments, shown in Figure~\ref{fig:monotonicity-pairs}, showed that while all of the quantifiers are learnable by the model, the monotone ones were reliably learned more quickly than the non-monotone one, suggesting that they are indeed easier to learn.

\begin{figure}[ht]
    \centering
    \includegraphics[width=\textwidth]{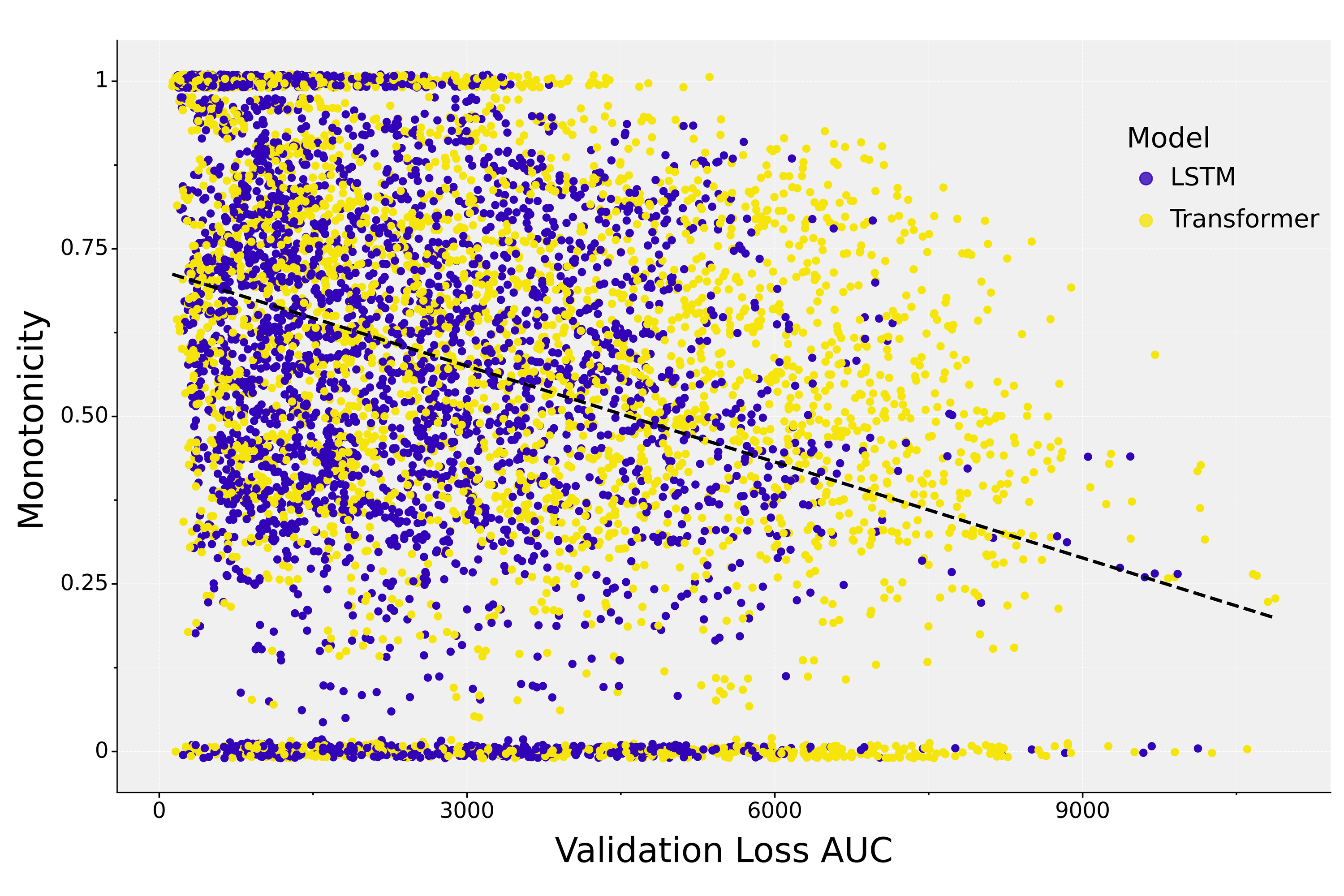}
    \caption{Results from \cite{haberlandQuantifiersThatAre2025}.  Each dot corresponds to one quantifier. The $x$-axis is the area under the learning curves; these are computed with respect to a loss, so lower values mean easier to learn.  The $y$-axis corresponds to the degree of monotonicity of each quantifier.  The line of best fit, in red, shows that there is a correlation between ease of learning and degree of monotonicity.}
    \label{fig:monotonicity-all}
\end{figure}

To more thoroughly develop the cognitive thermometer metaphor and build the aforementioned heatmap, \textcite{haberlandQuantifiersThatAre2025} expanded these experiments in two directions.  Firstly, they tested a wide range of quantifiers by sampling mathematical expressions for their denotations from a context-free grammar.  Secondly, they generalized a definition of a \emph{degree} of monotonicity \parencite{carcassiMonotoneQuantifiersEmerge2021b} as the mutual information between any function and its so-defined minimal monotone extension.  Figure~\ref{fig:monotonicity-all} shows that this degree of monotonicity is well correlated with a standard measure of ease of learning: quantifiers that are more monotone are easier to learn.

To summarize: we have used machine learning as a tool to measure the ease of learning of various semantic objects in the service of explaining linguistic universals.  Similar patterns have been found for other quantifier universals, as well as for convexity of color terms \parencite{steinert-threlkeldEaseLearningExplains2020} and certain attitude predicates \parencite{steinert-threlkeldExplanationVeridicalUniformity2020}.  We think of these machine learning models as cognitive thermometers because they are a measurement device that lets us build a heatmap across large semantic spaces.

\section{(Machine) Learning and (Logical) Complexity}
\label{sec:complexity}

While we have motivated the use of machine learning as a cognitive thermometer due to its ability to measure ease of learning at scale without language-specific confounds, it may not be the only useful device for measuring the target that we are after.  A particularly natural reaction would argue that \emph{simplicity} plays a key role: perhaps the universally expressed meanings are just simpler. We explore this thought by looking at several measures of simplicity and how they interact with the relevant universals. Our conclusion will be that (i) each such measure earns its explanatory power only by first fixing a representation language, a set of primitive operations, or an encoding, whereas the learnability thermometer commits to no such thing.  Moreover, (ii) in many cases, we will see that learning has more explanatory power than complexity.  Because of these points, we argue that ease of learning is not merely one measurement device among others, but is conceptually prior to the complexity measures.

First, following a prominent line in cognitive science, we investigate logical complexity \parencite{feldmanMinimizationBooleanComplexity2000, goodmanRationalAnalysisRulebased2008, goodmanConceptsProbabilisticLanguage2015}.\footnote{Many of these papers show empirically that human concept learning tracks logical complexity.  The Bayesian approach to cognitive science can be viewed as building this link between learning and simplicity in to the theory: when using structured representations, the prior is heavily biased towards simple representations \parencite{griffithsBayesianModelsCognition2024, griffithsBayesianModelsCognition2024a, quilty-dunnBestGameTown2023}.}  Roughly: given a plausible representation language for meanings, take the simplicity of a meaning to be the length of the shortest expression in that logical language which denotes it.  So, for instance, `every A is B' might be represented by $A \subseteq B$, but not by the equivalent $\lnot\left(B \setminus A = \emptyset \right)$ because the latter is more complex.\footnote{\textcite{walshGeneralizedQuantifiersNatural2026} has recently argued that the generalized quantifiers in natural language are all either $\Delta^1_1$ or $\Sigma^1_1$-definable, i.e.\ are of relatively low descriptive complexity broadly. This complements rather than competes with the present program: Walsh assumes two universals (conservativity and domain-independence) to derive a complexity ceiling, so his bound is downstream of the universals, whereas we treat the universals as the explanandum. His result also bounds the class as a whole without distinguishing between the monotone and non-monotone quantifiers within it, leaving open whether ease of learning refines the bound by recovering that finer structure.}
\Textcite{polQuantifiersSatisfyingSemantic2023} analyzed quantifiers from this perspective, and found that monotone quantifiers do tend to have shorter minimum description lengths than non-monotone ones, suggesting that learning and logical complexity broadly agree in this case.
In a follow-up analysis of the learning data from \cite{haberlandQuantifiersThatAre2025}, we performed a regression comparing logical complexity and ease of learning as predictors of the degree of monotonicity.  The results are clear: ease of learning has a much stronger effect than complexity. When added as a predictor with complexity, learning explains significantly more variance in monotonicity.  When complexity is added as a predictor over and above ease of learning, the additional explained variance is very small.  This suggests that ease of learning plays a stronger role than complexity in explaining monotonicity.

\begin{figure}[ht]
  \centering
  \includegraphics[width=\linewidth]{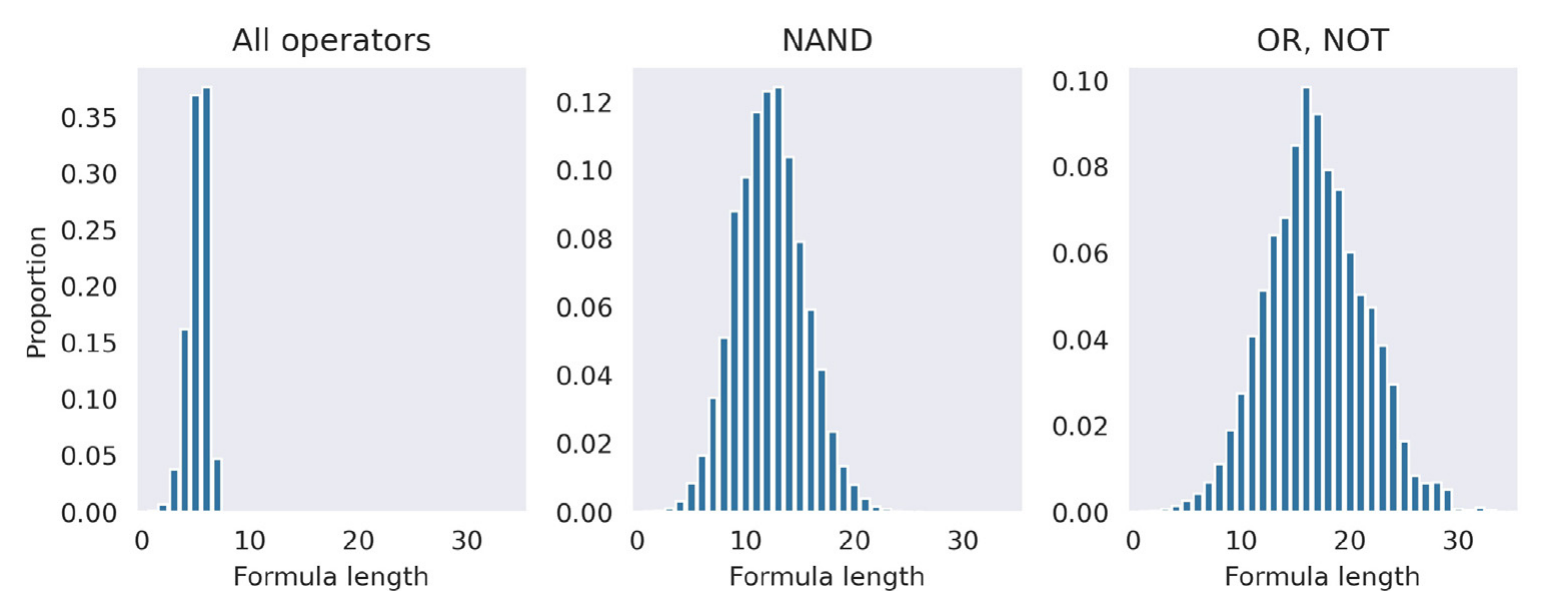}
  \caption{The complexity of a Boolean meaning is defined only relative to a
  choice of primitive operators. Distribution of minimal-formula lengths across
  all $65{,}536$ Boolean categories under three languages of thought: all nine
  operators (left), \textsc{nand} alone (centre), and $\{\textsc{or},\textsc{not}\}$
  (right). The same meanings that the nine-operator language expresses in at
  most eight symbols require up to sixteen operators under \textsc{nand} and up
  to thirty-five under $\{\textsc{or},\textsc{not}\}$. Adapted from
  \citet{CARCASSI2023105541}.}
  \label{fig:operator-dependence}
\end{figure}

\textcite{carcassiNeuralNetworksTrack2022} show that neural-network learning effort and logical complexity are positively rank-correlated across Boolean categories, but that the strength of the correlation depends on which operators the language of thought admits. This is an example of language-relativity of complexity. The minimal description length is defined only relative to a set of primitive operators, and different choices of that set induce very different complexity profiles. The same 65,536 categories that a nine-operator language expresses in at most eight symbols require up to sixteen under a \textsc{nand}-only language and up to thirty-five under $\{\textsc{or},\textsc{not}\}$ (Figure~\ref{fig:operator-dependence}). Which meanings count as ``simple''---and hence, on the simplicity story, easy to learn---thus depends on a prior choice of language.

The problem is wider, and one cannot escape it by reading the operators off behaviour: \textcite{CARCASSI2023105541} find that the true Boolean language of thought (LoT) is recoverable from category-learning data only in narrowly restricted regimes---few operators, a strong simplicity bias, and a favourable design---succeeding in well under one percent of their simulated experiments. In general, then, both the complexity of a meaning and the language relative to which it is defined are under-determined by the learning data. Machine learning is attractive precisely because it forces neither choice up front.

A candidate complexity measure that purports to overcome this language-dependence is Kolmogorov complexity: for any object $x$ (encoded as a binary sequence), $K(x)$ is the length of the shortest program for a universal Turing machine to compute $x$ \parencite{liIntroductionKolmogorovComplexity2008}. A famous invariance result states that the choice of universal TM (roughly, choice of description language) only affects $K(x)$ by an additive constant that depends only on the two TMs, but not the object $x$.  \Textcite{polQuantifiersSatisfyingSemantic2023} measured quantifier complexity using an empirical approximation to the uncomputable $K(x)$ (Lempel-Ziv complexity) and found that, while monotone quantifiers are less complex, quantifiers satisfying other universals are often more complex on this measure.  Furthermore, the language-invariance of $K(x)$ does not obviate the need for modeling choices in a cognitive context \parencite{chaterSimplicityUnifyingPrinciple2003}: the additive constant merely relocates the arbitrary choice from the description language to the encoding of objects by the mind, and this encoding strongly affects the empirical predictions made by this measure of simplicity.  For example: different ways of representing quantifiers as binary strings for computing $K(x)$ can yield different relative orderings of their complexity.

The sensitivity of both forms of complexity to language and representational choices points to another virtue of using machine learning in this application: while such learners do have their own inductive biases, the assumptions made to apply such models are more domain-general than the assumptions needed for other complexity measures. Crucially, this domain-generality is not only an argument we can make theoretically. Because the learnability ordering we measure is reproduced across markedly different architectures---LSTMs and Transformers alike (Figure~\ref{fig:monotonicity-all})---we have direct evidence that it does not simply reflect the inductive bias of one particular learner. Where a complexity measure's verdict can be flipped by re-choosing its primitives, the ease-of-learning verdict has so far proved stable under analogous alternative choices. That stability licenses treating ease of learning as a measurement of the meanings themselves, rather than an artifact of the instrument. How far this learner-invariance extends is itself an important question, to which we return below.

\section{Moving Forward}
\label{sec:moving-forward}

We have used machine learning to measure the temperature of meaning spaces and found semantic universals in warm regions. We close by asking what must be true of the world for this to work. To that end, we also offer a conjecture which captures the assumption on which the program rests.

We have spoken of ``ease of learning'' as a property of a meaning. Yet every learner brings its own inductive biases, and one might worry that our heat-map is an artifact of the instrument: other learning architectures may produce different warm regions. The evidence gives reason for optimism: the ordering of quantifiers by ease of learning is largely preserved across learners as different as LSTMs and transformers. We propose to take this as evidence for a more general claim:

\begin{description}
    \item[\textsc{Ease-of-Learning Invariance Thesis:}] There is an ordering of meanings by ease of learning that is invariant across sufficiently general learners---up to the learning analog of an additive constant---and it is this ordering that natural language semantic universals track.
\end{description}

The analogy is with Kolmogorov complexity, where the choice of universal machine shifts $K(x)$ only by a constant independent of $x$; here it is the choice of learner, not of description language, that is meant to wash out not the absolute effort of acquisition, which surely varies, but the order it imposes on the space of meanings. In that light, the previous section's moral becomes positive: each symbolic measure of simplicity fixes an arbitrary language and reads complexity off it, whereas ease of learning is the invariant those measures were approximating, agreeing with it when their primitives happen to be well-adapted, and diverging otherwise.

Unqualified, the thesis is false, and instructively so. No-free-lunch results \parencite{wolpertLackPrioriDistinctions1996, wolpertNoFreeLunch1997} guarantee that no learner is best for every target: over wholly unnatural spaces, different learning biases give different orders. All the content lives in sufficiently natural spaces, which is the bet that cognitively plausible, domain-general learners converge on a single order over the meanings that human languages actually use. Delimiting that class, and the domains on which convergence holds, is not a caveat but the program's central problem.

If the bet is right, the payoff runs both ways. Forward, a single learnability scale would organize universals well beyond quantifiers---across color, kinship, modality, and more---a common temperature we would expect to coincide with pressures such as communicative efficiency. Backward, the typological record becomes an instrument of its own: from the regions the world's languages show us are warm, we can begin to reverse-engineer the human learner's prior. The thermometer is always only an instrument. Future research will uncover the thermodynamics.

\printbibliography

\end{document}